\documentclass[10pt,twocolumn,letterpaper]{article}

\usepackage{cvpr}              

\usepackage{graphicx}
\usepackage{amsmath}
\usepackage{amssymb}
\usepackage{booktabs}

\usepackage{url}            
\usepackage{amsfonts}       
\usepackage{nicefrac}       
\usepackage{microtype}      
\usepackage{enumitem}
\usepackage{amsmath,amssymb,amsbsy}
\usepackage{multirow}
\usepackage{multicol}
\usepackage{wrapfig}
\usepackage{algorithm}
\usepackage{algpseudocode}
\usepackage{caption}
\usepackage{subcaption}
\usepackage{xcolor}

\usepackage[pagebackref,breaklinks,colorlinks]{hyperref}

\usepackage[capitalize]{cleveref}
\crefname{section}{Sec.}{Secs.}
\Crefname{section}{Section}{Sections}
\Crefname{table}{Table}{Tables}
\crefname{table}{Tab.}{Tabs.}

\def\cvprPaperID{5093} 
\def\confName{CVPR}
\def\confYear{2022}

\begin{document}

\title{CMT: Convolutional Neural Networks Meet Vision Transformers}

\author{Jianyuan Guo$^{1,2}$, Kai Han$^{2}$, Han Wu$^{1}$, Yehui Tang$^{2}$, Xinghao Chen$^{2}$, Yunhe Wang$^{2}$\thanks{Corresponding author. Mindspore~\cite{mindspore} implementation can be found in: \href{https://gitee.com/mindspore/models/tree/master/research/cv/CMT}{https://gitee.com/mindspore/models/tree/master/research/cv/CMT}. Pytorch~\cite{pytorch} implementation: \href{https://github.com/ggjy/CMT.pytorch}{https://github.com/ggjy/CMT.pytorch}.}, Chang Xu$^{1*}$ \\
	\normalsize$^1$ School of Computer Science, Faculty of Engineering, University of Sydney.
	\normalsize$^2$ Huawei Noah’s Ark Lab.
	\\
	\small\texttt{\{jianyuan.guo, kai.han, yunhe.wang\}@huawei.com; c.xu@sydney.edu.au}
}

\maketitle

\begin{abstract}
Vision transformers have been successfully applied to image recognition tasks due to their ability to capture long-range dependencies within an image. However, there are still gaps in both performance and computational cost between transformers and existing convolutional neural networks (CNNs). In this paper, we aim to address this issue and develop a network that can outperform not only the canonical transformers, but also the high-performance convolutional models. We propose a new transformer based hybrid network by taking advantage of transformers to capture long-range dependencies, and of CNNs to extract local information. Furthermore, we scale it to obtain a family of models, called CMTs, obtaining much better trade-off for accuracy and efficiency than previous CNN-based and transformer-based models. In particular, our CMT-S achieves 83.5\% top-1 accuracy on ImageNet, while being 14x and 2x smaller on FLOPs than the existing DeiT and EfficientNet, respectively. The proposed CMT-S also generalizes well on CIFAR10 (99.2\%), CIFAR100 (91.7\%), Flowers (98.7\%), and other challenging vision datasets such as COCO (44.3\% mAP), with considerably less computational cost.
\end{abstract}

\section{Introduction}
The past decades have witnessed the extraordinary contribution of CNNs~\cite{resnet,vggnet,mobilenetv2,inceptionv3,efficientnet} in the field of computer vision due to its ability of extracting deep discriminative features. 
Meanwhile, self-attention based transformers~\cite{bert,attention} has become the de facto most popular models for natural language processing (NLP) tasks, and shown excellent capability of capturing long-distance relationships. 
Recently, many researchers attempt to apply the transformer-based architectures to vision domains, and achieve promising results in various tasks such as image classification~\cite{vit,deit}, object detection~\cite{detr, deformable-detr}, and semantic segmentation~\cite{setr}. 
Vision transformer (ViT)~\cite{vit} is the first work to replace the conventional CNN backbone with a pure transformer. Input images (224$\times$224$\times$3) are first split into 196 non-overlapping patches (with a fixed size of 16$\times$16$\times$3 per patch), which are analogous to the word tokens in NLP. The patches are then fed into stacked standard transformer blocks to model global relations and extract feature for classification. The design paradigm of ViT has heavily inspired the following transformer based models for computer vision, such as IPT~\cite{ipt} for low-level vision and SETR~\cite{setr} for semantic segmentation.

Despite that transformers have demonstrated excellent capabilities when migrated to vision tasks, their performances are still far inferior to similar-sized convolutional neural network counterparts, \eg, EfficientNets~\cite{efficientnet}. 
We believe the reason of such weakness is threefold. Firstly, images are split into patches in ViT~\cite{vit} and other transformer-based models such as IPT~\cite{ipt} and SETR~\cite{setr}. Doing so can greatly simplify the process of applying transformer to image-based tasks. And the sequence of patches can be directly fed into a standard transformer where long-range dependencies between patches can be well captured. However, it ignores the fundamental difference between sequence-based NLP tasks and image-based vision tasks, \eg, the 2D structure and spatial local information within each patch.
Secondly, transformer is difficult to explicitly extract low-resolution and multi-scale features due to the fixed patch size, which poses a big challenge to dense prediction tasks such as detection and segmentation.
Thirdly, the computational and memory cost of self-attention modules in transformers are quadratic ($\mathcal{O}(N^2C)$) to the resolution of inputs, compared to $\mathcal{O}(NC^2)$ of convolution-based CNNs. High resolution images are prevalent and common, \eg, 1333$\times$800 in COCO~\cite{coco} and 2048$\times$1024 in Cityscapes~\cite{cityscapes}. Using transformers to process such images would inevitably cause the problem of insufficient GPU memory and low computation efficiency.

\begin{figure*}
	\centering
	\begin{subfigure}[b]{\columnwidth}
		\centering
		\includegraphics[width=\columnwidth]{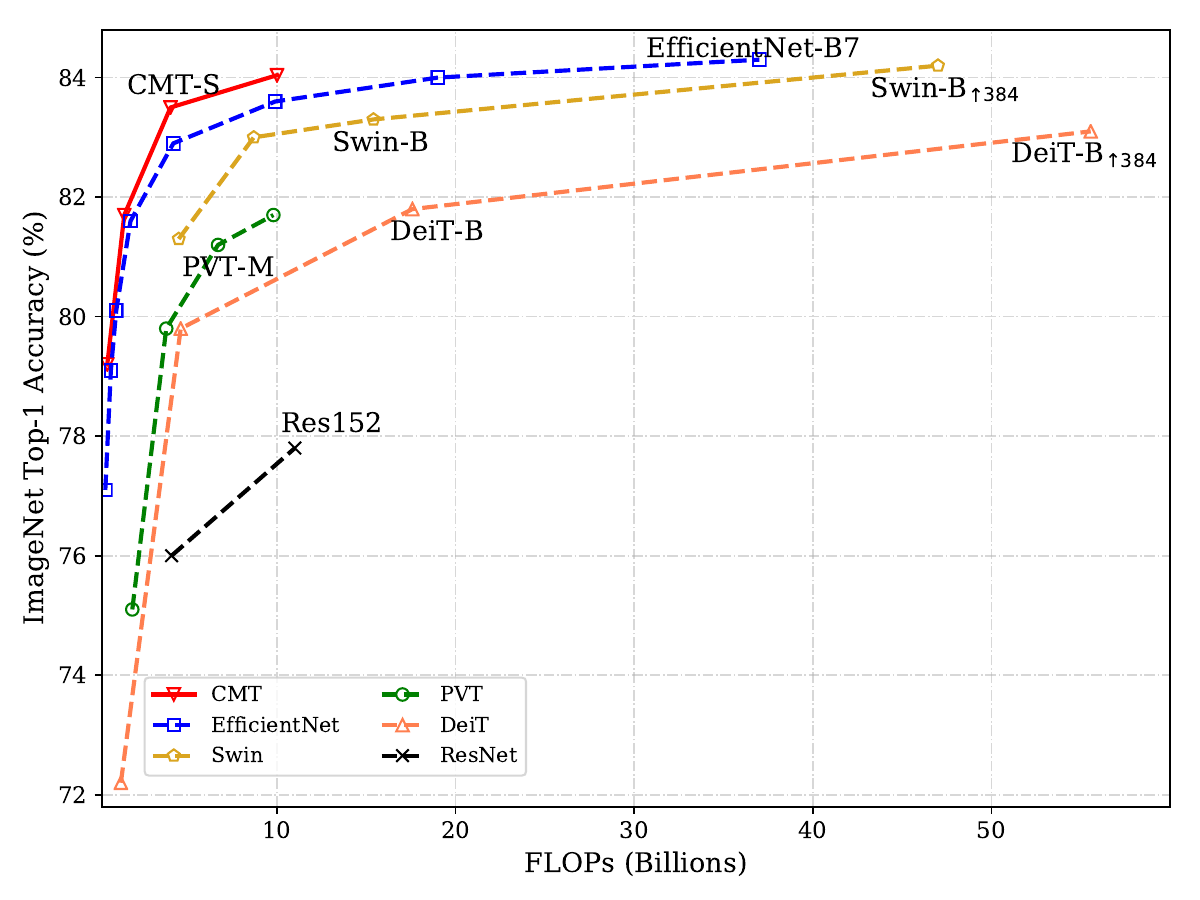}
		\hspace{-44mm}\resizebox{.48\columnwidth}{!}{
			\begin{tabular}[b]{l|ccc}
				\footnotesize
				Model & Top1 Acc. & \# Params & \# FLOPs \\
				\hline
				DeiT-Ti~\cite{deit} & 72.2\% & \bf 5M & 1.3B \\
				CPVT-Ti-GAP~\cite{cpvt} & 74.9\% & 6M & 1.3B \\
				CeiT-T~\cite{ceit} & 76.4\% & 5M & \bf 1.2B \\
				EfficientNet-B3~\cite{efficientnet} & 81.6\% & 12M & 1.8B \\
				\bf CMT-XS & \bf 81.7\% & 14M & 1.5B \\
				\hline
				ResNet-50~\cite{resnet} & 76.2\% & 26M & 4.1B \\
				DeiT-S~\cite{deit} & 79.8\% & 22M & 4.6B \\
				RegNetY-4GF~\cite{regnety} & 80.0\% & 21M & 4.0B \\
				T2T-ViT-14~\cite{t2t} & 80.6\% & 21M & 4.8B \\
				ResNeXt-101~\cite{resnext} & 80.9\% & 84M & 32B \\
				PVT-M~\cite{pvt} & 81.2\% & 44M & 6.7B \\
				Swin-T~\cite{swin} & 81.3\% & 29M & 4.5B \\
				CPVT-S-GAP~\cite{cpvt} & 81.5\% & 23M & 4.6B \\
				CeiT-S~\cite{ceit} & 82.0\% & 24M & 4.5B \\
				CvT-13-NAS~\cite{cvt} & 82.2\% & \bf 18M & 4.1B \\
				EfficientNet-B4~\cite{efficientnet} & 82.9\% & 19M & 4.2B \\
				\bf CMT-S & \bf 83.5\%  & 25M & \bf 4.0B \\ \hline
				\multicolumn{4}{c}{\vspace{15mm}}
		\end{tabular}}
		\vspace{-0.2cm}
		\caption{\small{ImageNet Accuracy vs. FLOPs}}
	\end{subfigure}
	\hfill
	\begin{subfigure}[b]{\columnwidth}
		\centering
		\includegraphics[width=\columnwidth]{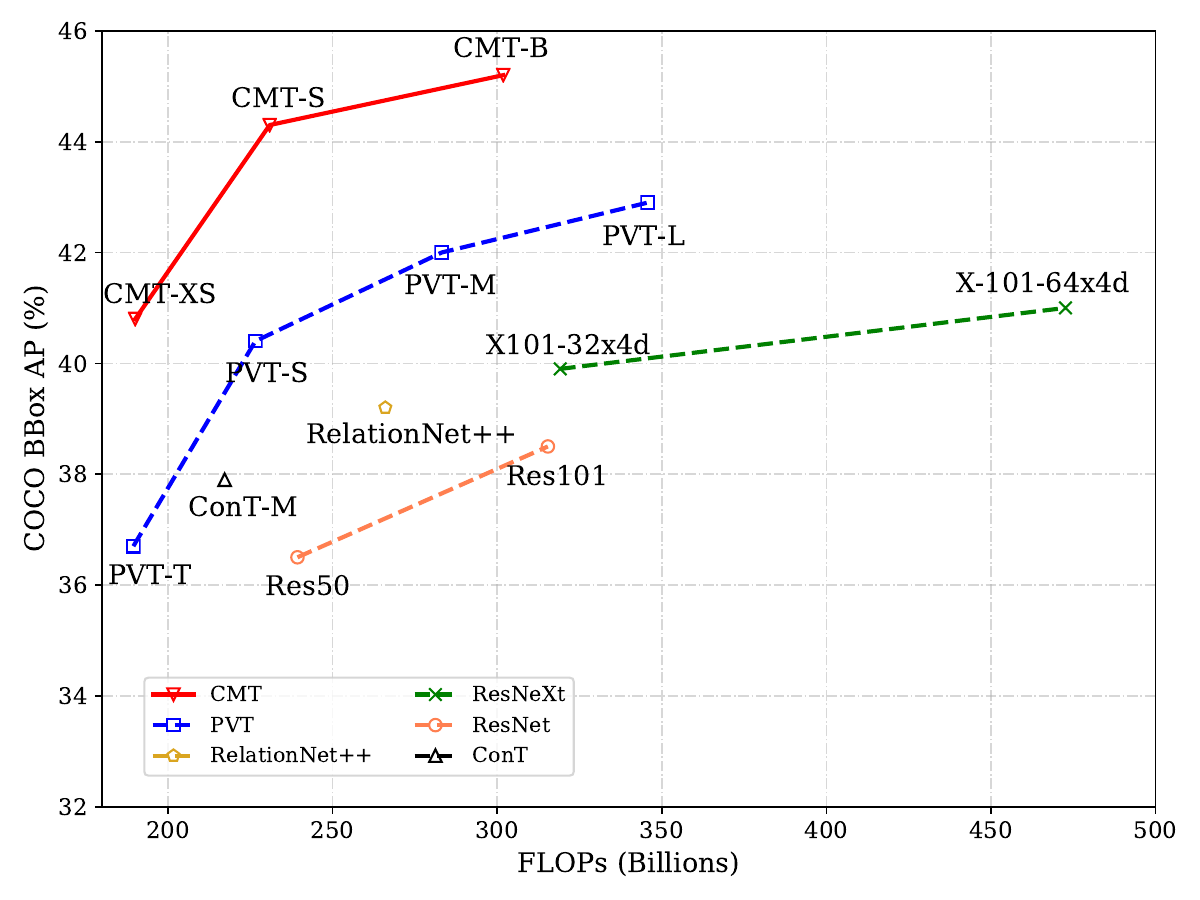}
		\hspace{-40mm}\resizebox{.42\columnwidth}{!}{
			\begin{tabular}[b]{l|cc}
				\footnotesize
				Model & mAP & \# FLOPs \\
				\hline
				ConT-M~\cite{cont} & 37.9\% & 217B \\
				ResNet-101~\cite{resnet} & 38.5\% & 315B \\
				RelationNet++~\cite{relationnet++} & 39.2\% & 266B \\
				ResNeXt-101-64x4d~\cite{resnext} & 41.0\% & 473B \\
				Swin-T~\cite{swin} & 41.5\% & 245B \\
				PVT-M~\cite{pvt} & 42.0\% & 283B \\
				Twins-SVT-S~\cite{twins} & 42.3\% & 209B \\
				\bf CMT-S & \bf 44.3\% & 230B \\ \hline
				\multicolumn{3}{c}{\vspace{15mm}}
		\end{tabular}}
		\vspace{-0.2cm}
		\caption{\small{COCO mAP vs. FLOPs}}
	\end{subfigure}
	\vspace{-0.3cm}
	\caption{\small{\textbf{Performance comparison between CMT and other models.} (a) Top-1 accuracy on ImageNet~\cite{imagenet}. (b) Object detection results on COCO val2017~\cite{coco} of different backbones using RetinaNet framework, all numbers are for single-scale, ``1x'' training schedule.}}
	\vspace{-0.4cm}
	\label{fig:imagenet-coco}
\end{figure*}

\begin{figure*}
    \centering         
	\includegraphics[width=\textwidth]{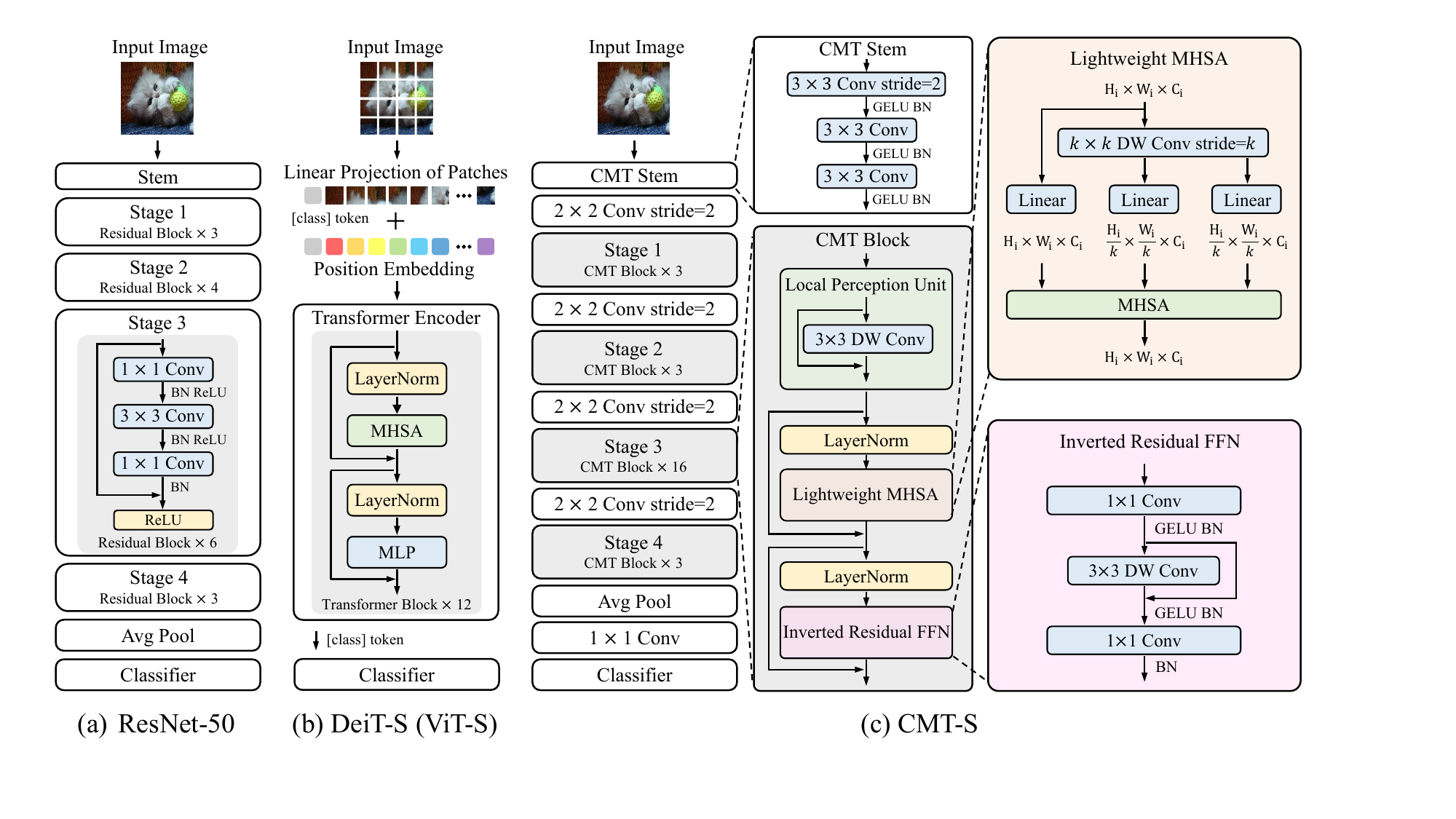}
	\vspace{-0.6cm}
	\caption{\small\textbf{Example of the CMT-S architecture.} (a) ResNet-50~\cite{resnet}. (b) DeiT-S~\cite{deit} (ViT-S~\cite{vit}) architecture, where MHSA denotes the multi-head self-attention module. (c) The proposed CMT-S, described in Sec.~\ref{sec:cmt}. More details and other variants are shown in Table~\ref{tab:arch}.}
	\vspace{-0.4cm}
	\label{fig:arch}
\end{figure*}

In this paper, we stand upon the intersection of CNNs and transformers, and propose a novel CMT (CNNs meet transformers) architecture for visual recognition. The proposed CMT takes the advantages of CNNs to compensate for the aforementioned limitations when utilizing pure transformers. As shown in Figure~\ref{fig:arch}(c), input images first go through the convolution stem for fine-grained feature extraction, and are then fed into a stack of CMT blocks for representation learning. Specifically, the introduced CMT block is an improved variant of transformer block whose local information is enhanced by depth-wise convolution. Compared to ViT~\cite{vit}, the features generated from the first stage of CMT can maintain higher resolution, \ie, $H/4\times W/4$ against $H/16\times W/16$ in ViT, which are essential for other dense prediction tasks. Furthermore, we adopt the stage-wise architecture design similar to CNNs~\cite{resnet,efficientnet,mobilenetv2} by using four convolutional layer with stride $2$, to gradually reduce the resolution (sequence length) and increase the dimension flexibly. The stage-wise design helps to extract multi-scale features and alleviate the computation burden caused by high resolution. The local perception unit (LPU) and inverted residual feed-forward network (IRFFN) in CMT block can help capture both local and global structure information within the intermediate features and promote the representation ability of the network. Finally, the average pooling is used to replace the class token in ViT for better classification results. In addition, we propose a simple scaling strategy to obtain a family of CMT variants. Extensive experiments on ImageNet and other downstream tasks demonstrate the superiority of our CMT in terms of accuracy and FLOPs. For example, our CMT-S achieves 83.5\% ImageNet top-1 with only 4.0B FLOPs, while being 14x and 2x less than the best existing DeiT~\cite{deit} and EfficientNet~\cite{efficientnet}, respectively. In addition to image classification, CMT can also be easily transferred to other vision tasks and serve as a versatile backbone. Using CMT-S as the backbone, RetinaNet~\cite{focal} can achieve 44.3\% mAP on COCO val2017, outperforming the PVT-based RetinaNet~\cite{pvt} by 3.9\% with less computational cost.

\section{Related Work}
The computer vision community prospered in past decades riding the wave of deep learning, and the most popular deep neural networks are often built upon basic blocks, in which a series of convolutional layers are stacked sequentially to capture local information within intermediate features. However, the limited receptive field of small convolutional kernels makes it difficult to obtain global information, withholding the networks of high performance on challenging tasks such as classification, object detection, and semantic segmentation. Therefore, many researchers start to dig deeper into self-attention based transformers which have the ability to capture long-range information. Here we briefly review the conventional CNNs and recently proposed vision transformers.

\vspace{0.1cm}
\noindent\textbf{Convolutional neural networks}.
The first standard CNN was proposed by LeCun~\etal~\cite{lenet} for handwritten character recognition, and the past decades have witnessed that many powerful networks~\cite{alexnet,googlenet,vggnet,resnet,densenet} achieved unprecedented success on large scale image classification task~\cite{imagenet}. AlexNet~\cite{alexnet} and VGGNet~\cite{vggnet} showed that a deep neural network composed of convolutional layers and pooling layers can obtain adequate results in recognition. GoogleNet~\cite{googlenet} and InceptionNet~\cite{inceptionv3} demonstrated the effectiveness of multiple paths within a basic block. ResNet~\cite{resnet} showed better generalization by adding shortcut connections every two layers to the base network. To alleviate the limited receptive fields in prior research, some researches~\cite{wang2017residual,cbam,senet,genet,bam,sasa} incorporated attention mechanisms as an operator for adaptation between modalities. Wang \etal~\cite{wang2017residual} proposed to stack attention modules sequentially between the intermediate stages of deep residual networks. SENet~\cite{senet} and GENet~\cite{genet} adaptively recalibrated
channel-wise feature responses by modeling interdependencies between channels. NLNet~\cite{nlnet} incorporated the self-attention mechanism into neural networks, providing pairwise interactions across all spatial positions to augment the long-range dependencies. In addition to above architectural advances, there has also been works~\cite{squeezenet,shufflenetv2,ghostnet} focusing on improving over-parameterized deep neural networks by trading accuracy for efficiency. For example, MobileNets~\cite{mobilenetv2,mobilenetv3} and EfficientNets~\cite{efficientnet} both leveraged neural architecture search (NAS) to design efficient mobile-size network and achieved new state-of-the-art results.

\vspace{0.1cm}
\noindent\textbf{Vision transformers.}
Since transformers achieved remarkable success in natural language processing (NLP)~\cite{attention,bert}, many attempts~\cite{vit,deit,li2022brain,pvt,cvt,hire,wave,cpvt,tnt,ceit,swin,tang2021patch} have been made to introduce transformer-like architectures to vision tasks. The pioneering work ViT~\cite{vit} directly applied the transformer architecture inherited from NLP to classification with image patches as input. While ViT required a large private dataset JFT-300M~\cite{jft} to achieve promising result, DeiT~\cite{deit} introduced a new training paradigm to extend ViT to a data-efficient transformer directly trained on ImageNet-1K. T2T-ViT~\cite{t2t} proposed to embed visual tokens by recursively aggregating neighboring tokens into one token. TNT~\cite{t2t} proposed to model both patch-level and pixel-level representation by the inner and outer transformer block, respectively. PVT~\cite{pvt} introduced the pyramid structure into ViT, which can generate multi-scale feature maps for various pixel-level dense prediction tasks. CPVT~\cite{cpvt} and CvT~\cite{cvt} are the most related to our work which leverage a convolutional projection into conventional transformer block, but we carefully investigate how to maximize the advantage of utilizing both CNNs and transformers by studying the different components including shortcut and normalization functions and successfully obtain a more superior result. Besides, transformers are also used to solve other vision tasks such as object detection~\cite{detr,deformable-detr}, semantic segmentation~\cite{setr}, image retrieval~\cite{retrieval}, and low-level vision task~\cite{ipt}.

Although there are many works successfully applying transformers for vision tasks, they have not shown satisfactory results compared to conventional CNNs, which are still the primary architectures for vision applications. Transformers are especially good at modeling long-range dependencies necessary for downstream vision tasks. However, locality should also be maintained for visual perception. In this paper, we demonstrate the potential of combining the transformer based network together with convolutional layer, the overall architecture follows the elaborated prior convolutional neural networks such as ResNet~\cite{resnet} and EfficientNet~\cite{efficientnet}.

\section{Approach}
\label{sec:cmt}
\subsection{Overall Architecture}
Our intention is to build a hybrid network taking the advantages of both CNNs and transformers. An overview of ResNet-50~\cite{resnet}, DeiT~\cite{deit}, and the proposed small version (CMT-S) of CMT architectures are presented in Figure~\ref{fig:arch}. As shown in Figure~\ref{fig:arch}(b), DeiT directly splits an input image into non-overlapping patches, however, the in-patch structure information can only be poorly modeled with linear projections. To overcome this limitation, we utilize the stem architecture~\cite{he2019bag} which has a $3\times3$ convolution with a stride of $2$ and an output channel of $32$ to reduce the size of input images, followed by another two $3\times3$ convolutions with stride $1$ for better local information extraction. Following the design in modern CNNs (\eg, ResNet~\cite{resnet}), our model has four stages to generate feature maps of different scales which are important for dense prediction tasks. To produce the hierarchical representation, a patch embedding layer consisting of a convolution and a layer normalization (LN)~\cite{ln} is applied before each stage to reduce the size of intermediate feature (2x downsampling of resolution), and project it to a larger dimension (2x enlargement of dimension). In each stage, several CMT blocks are stacked sequentially for feature transformation while retaining the same resolution of the input. For example, the ``Stage 3'' of CMT-S contains $16$ CMT blocks as illustrated in Figure~\ref{fig:arch}(c). The CMT block is able to capture both local and long-range dependencies, and we will describe it in Sec.~\ref{sec:cmt-block} in details. The model ends with a global average pooling layer, a projection layer, and a 1000-way classification layer with softmax.

Given an input image, we can obtain four hierarchical feature maps with different resolutions, similar to typical CNNs such as ResNet~\cite{resnet} and EfficientNet~\cite{efficientnet}. With the above feature maps whose strides are $4$, $8$, $16$, and $32$ with respect to the input, our CMT can obtain multi-scale representations of input images and can be easily applied to downstream tasks such as object detection and semantic segmentation.

\subsection{CMT Block}
\label{sec:cmt-block}
The proposed CMT block consists of a local perception unit (LPU), a lightweight multi-head self-attention (LMHSA) module, and an inverted residual feed-forward network (IRFFN), as illustrated in Figure~\ref{fig:arch}(c). We will describe these three parts in the following.

\vspace{0.1cm}
\noindent\textbf{Local Perception Unit.} Rotation and shift are two commonly used data augmentation manners in vision tasks, and these operations should not alter the final results of the model. In other words, we expect translation-invariance~\cite{translation-invariance} in those tasks.
However, the absolute positional encoding used in previous transformers, initially designed to leverage the order of tokens, damages such invariance because it adds unique positional encoding to each patch~\cite{cpvt}.
Besides, vision transformers ignore the local relation~\cite{local-scale} and the structure information~\cite{zero-padding} inside the patch. To alleviate the limitations, we propose the local perception unit (LPU) to extract local information, which is defined as:
\begin{equation}
\mathrm{LPU}(\mathrm{\bf X}) = \mathrm{DWConv}(\mathrm{\bf X}) + \mathrm{\bf X}.
\end{equation}
where $\mathrm{\bf X}\in\mathbb{R}^{H\times W\times d}$, $H\times W$ is the resolution of the input of current stage, $d$ indicates the dimension of features. $\mathrm{DWConv}(\cdot)$ denotes the depth-wise convolution.

\vspace{0.1cm}
\noindent\textbf{Lightweight Multi-head Self-attention.} In original self-attention module, the input $\mathrm{\bf X}\in\mathbb{R}^{n\times d}$ is linearly transformed into query $\mathrm{\bf Q}\in\mathbb{R}^{n\times d_k}$, key $\mathrm{\bf K}\in\mathbb{R}^{n\times d_k}$, and value $\mathrm{\bf V}\in\mathbb{R}^{n\times d_v}$, where $n=H\times W$ is the number of patches. And we omit the reshape operation from $ H\times W\times d$ to $n\times d$ of tensors in Figure~\ref{fig:arch}(c) for simplicity. The notation $d$, $d_k$ and $d_v$ are the dimensions of input, key (query) and value, respectively. Then the self-attention module is applied as:
\begin{equation}
\mathrm{Attn}(\textbf{Q}, \textbf{K}, \textbf{V}) = \mathrm{Softmax}(\frac{\textbf{Q}\textbf{K}^T}{\sqrt{d_k}})\textbf{V}.
\end{equation}
To mitigate the computation overhead, we use a $k\times k$ depth-wise convolution with stride $k$ to reduce the spatial size of $\mathrm{\bf K}$ and $\mathrm{\bf V}$ before the attention operation, \ie, $\mathrm{\bf K'}=\mathrm{DWConv}(\mathrm{\bf K})\in\mathbb{R}^{\frac{n}{k^2}\times d_k}$ and $\mathrm{\bf V'}=\mathrm{DWConv}(\mathrm{\bf V})\in\mathbb{R}^{\frac{n}{k^2}\times d_v}$ as shown in Figure~\ref{fig:arch}(c). In addition, we add a relative position bias $\mathrm{\bf B}$ to each self-attention module, and the corresponding lightweight attention is defined as:
\begin{equation}
\mathrm{LightAttn}(\textbf{Q}, \textbf{K}, \textbf{V}) = \mathrm{Softmax}(\frac{\textbf{Q}\textbf{K}'^{T}}{\sqrt{d_k}} + \textbf{B})\textbf{V}'.
\end{equation}
where $\mathrm{\bf B}\in\mathbb{R}^{n\times \frac{n}{k^2}}$ is randomly initialized and learnable. The learnt relative position bias can also be easily transferred to $\mathrm{\bf B'}\in\mathbb{R}^{m_1\times m_2}$ with a different size $m_1\times m_2$ through bicubic interpolation, \ie, $\mathrm{\bf B'}=\mathrm{Bicubic}(\mathrm{\bf B})$. Thus it is convenient to fine-tune the proposed CMT for other downstream vision tasks. Finally, the lightweight multi-head self-attention (LMHSA) module is defined by considering $h$ “heads”, \ie, $h$ LightweightAttention functions are applied to the input. Each head outputs a sequence of size $n \times \frac{d}{h}$. These $h$ sequences are then concatenated into a $n\times d$ sequence.

\vspace{0.1cm}
\noindent\textbf{Inverted Residual Feed-forward Network.} The original FFN proposed in ViT~\cite{vit} is composed of two linear layers  separated by a GELU activation~\cite{gelu}. The first layer expands the dimension by a factor of $4$, and the second layer reduces the dimension by the same ratio:
\begin{equation}
\mathrm{FFN}(\mathrm{\bf X}) = \mathrm{GELU}(\mathrm{\bf X}W_1 + b_1)W_2 + b_2.
\end{equation}
where $W_1\in\mathbb{R}^{d\times 4d}$ and $W_2\in\mathbb{R}^{4d\times d}$ indicate weights of the two linear layers, respectively. The notation $b_1$ and $b_2$ are the bias terms. Figure~\ref{fig:arch}(c) provides a schematic visualization of our design. The proposed inverted residual feed-forward network (IRFFN) appears similar to inverted residual block~\cite{mobilenetv2} consisting of an expansion layer followed by a depth-wise convolution and a projection layer. Specifically, we change the location of shortcut connection for better performance:
\begin{align}
\mathrm{IRFFN}(\mathrm{\bf X}) &= \mathrm{Conv}(\mathcal{F}(\mathrm{Conv}(\mathrm{\bf X}))),\\
\mathcal{F}(\mathrm{\bf X}) &= \mathrm{DWConv}(\mathrm{\bf X}) + \mathrm{\bf X}.
\end{align}
where the activation layer is omitted. We also include the batch normalization after the activation layer and the last linear layer according to \cite{mobilenetv2}. The depth-wise convolution is used to extract local information with negligible extra computational cost. The motivation for inserting shortcut is similar to that of classic residual networks, which can promote the propagation ability of gradient across layers. We show that such shortcut helps the network achieve better results in our experiments.

With the aforementioned three components, the CMT block can be formulated as:
\begin{align}
{\bf Y}_i &= \mathrm{LPU}({\bf X}_{i-1}), \\
{\bf Z}_i &= \mathrm{LMHSA}(\mathrm{LN}({\bf Y}_i)) + {\bf Y}_i, \\
{\bf X}_i &= \mathrm{IRFFN}(\mathrm{LN}({\bf Z}_i)) + {\bf Z}_i.
\end{align}
where ${\bf Y}_i$ and ${\bf Z}_i$ denote the output features of LPU and LMHSA module for the $i$-th block, respectively. LN denotes the layer normalization~\cite{ln}. We stack several CMT blocks in each stage for feature transformation and aggregation.

\begin{table*}
	\renewcommand{\arraystretch}{0.88}
	\setlength\tabcolsep{5pt}
	\caption{\small\textbf{Architectures for ImageNet classification.} The output size corresponds to the input resolution of $224\times224$. Convolutions and CMT blocks are shown in brackets with the number of stacked blocks (see also Figure~\ref{fig:arch}(c)). $H_i$ and $k_i$ are the number of heads and reduction rates in LMHSA of stage $i$, respectively. $R_i$ denotes the expansion ratio in IRFFN of stage $i$.}
	\vspace{-0.3cm}
	\centering
	\small
	\begin{tabular}{c|c|c|c|c|c}
		\toprule [0.15em]
		Output Size & Layer Name & CMT-Ti & CMT-XS & CMT-S & CMT-B \\ \midrule
		$112\times112$ & Stem
		& $\begin{array}{c} 3\times3, 16, \text{stride}~2 \\ \left[3\times3, 16\right] \times 2 \end{array}$
		& $\begin{array}{c} 3\times3, 16, \text{stride}~2 \\ \left[3\times3, 16\right] \times 2 \end{array}$
		& $\begin{array}{c} 3\times3, 32, \text{stride}~2 \\ \left[3\times3, 32\right] \times 2 \end{array}$
		& $\begin{array}{c} 3\times3, 38, \text{stride}~2 \\ \left[3\times3, 38\right] \times 2 \end{array}$ \\ \midrule
		
		$56\times56$ & Patch Embedding & $2\times2$, $46$, stride $2$ & $2\times2$, $52$, stride $2$ & $2\times2$, $64$, stride $2$ & $2\times2$, $76$, stride $2$ \\ \midrule
		
		\begin{tabular}{c} Stage 1 \end{tabular} & \begin{tabular}{c}LPU \\LMHSA\\ IRFFN\end{tabular}
		& $\begin{bmatrix}\setlength{\arraycolsep}{1pt} \begin{array}{c}
		3\times3, 46\\ H_1$=$1, k_1$=$8\\ R_1$=$3.6
		\end{array} \end{bmatrix} \times 2$
		& $\begin{bmatrix}\setlength{\arraycolsep}{1pt} \begin{array}{c}
		3\times3, 52\\ H_1$=$1, k_1$=$8\\ R_1$=$3.8
		\end{array} \end{bmatrix} \times 3$
		& $\begin{bmatrix}\setlength{\arraycolsep}{1pt} \begin{array}{c}
		3\times3, 64\\ H_1$=$1, k_1$=$8\\ R_1$=$4  
		\end{array} \end{bmatrix} \times 3$
		& $\begin{bmatrix}\setlength{\arraycolsep}{1pt} \begin{array}{c}
		3\times3, 76\\ H_1$=$1, k_1$=$8\\ R_1$=$4  
		\end{array} \end{bmatrix} \times 4$ \\ \midrule
		
		$28\times28$ & Patch Embedding & $2\times2$, $92$, stride $2$ & $2\times2$, $104$, stride $2$ & $2\times2$, $128$, stride $2$ & $2\times2$, $152$, stride $2$ \\ \midrule 
		\begin{tabular}{c} Stage 2 \end{tabular} & \begin{tabular}{c}LPU \\ LMHSA \\ IRFFN \end{tabular} 
		& $\begin{bmatrix}\setlength{\arraycolsep}{1pt} \begin{array}{c}
		3\times3, 92\\ H_2$=$2, k_2$=$4\\ R_2$=$3.6
		\end{array} \end{bmatrix} \times 2$
		& $\begin{bmatrix}\setlength{\arraycolsep}{1pt} \begin{array}{c}
		3\times3, 104\\ H_2$=$2, k_2$=$4\\ R_2$=$3.8
		\end{array} \end{bmatrix} \times 3$
		& $\begin{bmatrix}\setlength{\arraycolsep}{1pt} \begin{array}{c}
		3\times3, 128\\ H_2$=$2, k_2$=$4\\ R_2$=$4
		\end{array} \end{bmatrix} \times 3$
		& $\begin{bmatrix}\setlength{\arraycolsep}{1pt} \begin{array}{c}
		3\times3, 152\\ H_2$=$2, k_2$=$4\\ R_2$=$4 
		\end{array} \end{bmatrix} \times 4$ \\ \midrule
		
		$14\times14$ & Patch Embedding & $2\times2$, $184$, stride $2$ & $2\times2$, $208$, stride $2$ & $2\times2$, $256$, stride $2$ & $2\times2$, $304$, stride $2$ \\ \midrule
		
		\begin{tabular}{c} Stage 3 \end{tabular} & \begin{tabular}{c}LPU\\ LMHSA\\ IRFFN\end{tabular} 
		& $\begin{bmatrix}\setlength{\arraycolsep}{1pt} \begin{array}{c}
		3\times3, 184\\ H_3$=$4, k_3$=$2\\ R_3$=$3.6
		\end{array} \end{bmatrix} \times 10$
		& $\begin{bmatrix}\setlength{\arraycolsep}{1pt} \begin{array}{c}
		3\times3, 208\\ H_3$=$4, k_3$=$2\\ R_3$=$3.8
		\end{array} \end{bmatrix} \times 12$
		& $\begin{bmatrix}\setlength{\arraycolsep}{1pt} \begin{array}{c}
		3\times3, 256\\ H_3$=$4, k_3$=$2\\ R_3$=$4 
		\end{array} \end{bmatrix} \times 16$
		& $\begin{bmatrix}\setlength{\arraycolsep}{1pt} \begin{array}{c}
		3\times3, 304\\ H_3$=$4, k_3$=$2\\ R_3$=$4 
		\end{array} \end{bmatrix} \times 20$ \\ \midrule
		
		$7\times7$ & Patch Embedding & $2\times2$, $368$, stride $2$ & $2\times2$, $416$, stride $2$ & $2\times2$, $512$, stride $2$ & $2\times2$, $608$, stride $2$ \\ \midrule
		
		\begin{tabular}{c} Stage 4 \end{tabular} & \begin{tabular}{c}LPU \\ LMHSA\\ IRFFN\end{tabular} 
		& $\begin{bmatrix}\setlength{\arraycolsep}{1pt} \begin{array}{c}
		3\times3, 368\\ H_4$=$8, k_4$=$1\\ R_4$=$3.6
		\end{array} \end{bmatrix} \times 2$
		& $\begin{bmatrix}\setlength{\arraycolsep}{1pt} \begin{array}{c}
		3\times3, 416\\ H_4$=$8, k_4$=$1\\ R_4$=$3.8
		\end{array} \end{bmatrix} \times 3$
		& $\begin{bmatrix}\setlength{\arraycolsep}{1pt} \begin{array}{c}
		3\times3, 512\\ H_4$=$8, k_4$=$1\\ R_4$=$4 
		\end{array} \end{bmatrix} \times 3$
		& $\begin{bmatrix}\setlength{\arraycolsep}{1pt} \begin{array}{c}
		3\times3, 608\\ H_4$=$8, k_4$=$1\\ R_4$=$4 
		\end{array} \end{bmatrix} \times 4$ \\ \midrule
		
		$1\times1$ & Projection & \multicolumn{4}{c}{$1\times1$, $1280$} \\ \midrule
		$1\times1$ & Classifier & \multicolumn{4}{c}{Fully Connected Layer, $1000$} \\ \midrule
		\multicolumn{2}{c|}{\# Params} & $9.49$ M & $15.24$ M & $25.14$ M & $45.72$ M \\ \midrule
		\multicolumn{2}{c|}{\# FLOPs} & $0.64$ B & $1.54$ B & $4.04$ B & $9.33$ B \\
		\bottomrule[0.15em]
	\end{tabular}
	\vspace{-0.1cm}
	\label{tab:arch}
\end{table*}

\subsection{Complexity Analysis}
We analyze the computational cost between standard ViT~\cite{vit} and our CMT in this section. A standard transformer block consists of a MHSA module and a FFN. Given an input feature of size $n\times d$, the computational complexity (FLOPs) can be calculated as:
\begin{align}
\vspace{-0.2cm}
\mathcal{O}(\mathrm{MHSA}) &= 2nd(d_k + d_v) + n^2(d_k + d_v), \\
\mathcal{O}(\mathrm{FFN}) &= 2nd^2r,
\end{align}
where $r$ is the expansion ratio of FFN, $d_k$ and $d_v$ are dimensions of key and value, respectively. More specifically, ViT sets $d=d_k=d_v$ and $r=4$, the cost can be simplified as:
\begin{equation}
\begin{split}
\mathcal{O}(\mathrm{Transformer\;block}) &= \mathcal{O}(\mathrm{MHSA}) + \mathcal{O}(\mathrm{FFN}) \\
&= 12nd^2 + 2n^2d
\end{split}
\end{equation}
Under above setting, the FLOPs of CMT block is as follows:
\begin{align}
\mathcal{O}(\mathrm{LPU}) &= 9nd, \\
\mathcal{O}(\mathrm{LMHSA}) &= 2nd^2(1+1/k^2) + 2n^2d/k^2, \\
\mathcal{O}(\mathrm{IRFFN}) &= 8nd^2 + 36nd,
\end{align}
\begin{equation}
\begin{aligned}
\mathcal{O}(\mathrm{CMT\;block}) &= \mathcal{O}(\mathrm{LPU})+\mathcal{O}(\mathrm{LMHSA}) + \mathcal{O}(\mathrm{IRFFN}) \\
&= 10nd^2(1+0.2/k^2) + 2n^2d/k^2 + 45nd,
\end{aligned}
\label{eq:cmt-block-flops}
\end{equation}
where $k\geq1$ is the reduction ratio in LMHSA. Compared to standard transformer block, the CMT block is more friendly to computational cost, and is easier to process the feature map under higher resolution (larger $n$).

\subsection{Scaling Strategy}
Inspired by~\cite{efficientnet}, we propose a new compound scaling strategy suitable for transformer-based networks, which uses a compound coefficient $\phi$ to uniformly scale the number of layers (depth), dimensions, and input resolution in a principled way:
\begin{equation}
\begin{aligned}
\mathrm{depth}: \alpha^\phi, \quad \mathrm{dimension}: \beta^\phi, \quad \mathrm{resolution}: \gamma^\phi, \\
\label{eq:scaling}
\mathrm{s.t.} \quad \alpha \cdot \beta^{1.5} \cdot \gamma^2 \approx 2.5, \quad \alpha \geq 1, \beta \geq 1, \gamma \geq 1
\end{aligned}
\end{equation}
where $\alpha$, $\beta$, and $\gamma$ are constants determined by grid search to decide how to assign resources to network depth, dimension and input resolution, respectively. Intuitively, $\phi$ is the coefficient that controls how many more ($\phi\geq1$) or less ($\phi\leq-1$) resources are available for model scaling. Notably, the FLOPs of the proposed CMT block is approximately proportional\footnote{The precious proportion is associated with $n$ and $d$. For example, CMT-S has $n$=$3136\gg d$=$64$ in ``stage 1'' and $n$=$49\ll d$=$512$ in ``stage 4''. The above proportion can already generate good variants for CMT.} to $\alpha$, $\beta^{1.5}$, and $\gamma^2$ according to E.q. \ref{eq:cmt-block-flops}. And we constraint $\alpha \cdot \beta^{1.5} \cdot \gamma^2 \approx 2.5$ so that for a given new $\phi$, the total FLOPS will approximately increase by $2.5^{\phi}$. This will strike a balance between the increase of computational cost and performance gain.
In our experiments, we empirically set $\alpha$=1.2, $\beta$=1.3, and $\gamma$=1.15.

We build our model CMT-S to have similar model size and computation complexity with DeiT-S (ViT-S) and EfficientNet-B4. We also introduce CMT-Ti, CMT-XS and CMT-B according to the proposed scaling strategy. The input resolutions are $160^2$, $192^2$, $224^2$, and $256^2$ for all four models, respectively. The detailed architecture hyper-parameters are shown in Table~\ref{tab:arch}.

\begin{table*}
	\caption{\small{\textbf{ImageNet Results of CMT}. CNNs and transformers with similar accuracy are grouped together for comparison. The proposed CMTs consistently outperform other methods with less computational cost.}}
	\vspace{-0.3cm}
	\centering
	\small
	\begin{tabular}{l||cc||cc||ccc}
		\toprule [0.15em]
		Model &  Top-1 Acc. & Top-5 Acc. & Throughput & \# Params & Resolution & \# FLOPs & Ratio \\
		\midrule [0.1em]
		CPVT-Ti-GAP~\cite{cpvt} & 74.9\% & - & - & 6M & 224$^2$ & 1.3B & 2.6$\times$ \\
		DenseNet-169~\cite{densenet} & 76.2\% & 93.2\% & - & 14M & 224$^2$ & 3.5B & 7$\times$  \\
		EfficientNet-B1~\cite{efficientnet} & 79.1\% & 94.4\% & - & 7.8M & 240$^2$ & 0.7B & 1.2$\times$ \\
		\bf CMT-Ti  & \bf 79.1\% & \bf 94.5\% & 1323.5 & 9.5M & 160$^2$ & \bf 0.6B & \bf 1$\times$ \\
		\midrule
		ResNet-50~\cite{resnet} & 76.2\% & 92.9\% & - & 25.6M & 224$^2$ & 4.1B & 2.7$\times$ \\
		CoaT-Lite Mini~\cite{coat} & 78.9\% & - & - & 11M & 224$^2$ & 2.0B & 1.3$\times$ \\
		DeiT-S~\cite{deit} & 79.8\% & - & 940.4 & 22M & 224$^2$ & 4.6B & 3.1$\times$ \\
		EfficientNet-B3~\cite{efficientnet} & 81.6\% & 95.7\% & 732.1 & 12M & 300$^2$ & 1.8B & 1.2$\times$ \\
		\bf CMT-XS  & \bf 81.8\% & \bf 95.8\% & 857.4 & 15.2M & 192$^2$ & \bf 1.5B & \bf 1$\times$ \\
		\midrule
		ResNeXt-101-64x4d~\cite{resnext} & 80.9\% & 95.6\% & - & 84M & 224$^2$ & 32B & 8$\times$ \\
		T2T-ViT-19~\cite{t2t} & 81.2\% & - & - & 39.0M & 224$^2$ & 8.0B & 2$\times$ \\
		PVT-M~\cite{pvt} & 81.2\% & - & 528.1 & 44.2M & 224$^2$ & 6.7B & 1.7$\times$ \\
		Swin-T~\cite{swin} & 81.3\% & - & 755.2 & 29M & 224$^2$ & 4.5B & 1.1$\times$ \\
		CPVT-S-GAP~\cite{cpvt} & 81.5\% & - & - & 23M & 224$^2$ & 4.6B & 1.2$\times$ \\
		RegNetY-8GF~\cite{regnety} & 81.7\% & - & 591.6 & 39.2M & 224$^2$ & 8.0B & 2$\times$ \\
		CeiT-S~\cite{ceit} & 82.0\% & 95.9\% & - & 24.2M & 224$^2$ & 4.5B & 1.1$\times$ \\
		EfficientNet-B4~\cite{efficientnet} & 82.9\% & 96.4\% & 349.4 & 19M & 380$^2$ & 4.2B & 1$\times$ \\
		Twins-SVT-B~\cite{twins} & 83.1\% & - & - & 56.0M & 224$^2$ & 8.3B & 2.1$\times$ \\
		\bf CMT-S  & \bf 83.5\% & \bf 96.6\% & 562.5 & 25.1M & 224$^2$ & \bf 4.0B & \bf 1$\times$ \\
		\midrule
		ViT-B/16$_{\uparrow 384}$~\cite{vit} & 77.9\% & - & 85.9 & 85.8M & 384$^2$ & 77.9B & 8.4$\times$ \\
		TNT-B~\cite{tnt} & 82.8\% & 96.3\% & - & 65.6M & 224$^2$ & 14.1B & 1.5$\times$ \\
		DeiT-B$_{\uparrow 384}$~\cite{deit} & 83.1\% & - & 85.9 & 85.8M & 384$^2$ & 55.6B & 6.0$\times$ \\
		CvT-21$_{\uparrow 384}$~\cite{cvt} & 83.3\% & - & - & 31.5M & 384$^2$ & 24.9B & 2.7$\times$ \\
		Swin-B~\cite{swin} & 83.3\% & - & 278.1 & 88M & 224$^2$ & 15.4B & 1.5$\times$ \\
		Twins-SVT-L~\cite{twins} & 83.3\% & - & 288.0 & 99.2M & 224$^2$ & 14.8B & 1.7$\times$ \\ 
		CeiT-S$_{\uparrow 384}$~\cite{ceit} & 83.3\% & 96.5\% & - & 24.2M & 384$^2$ & 12.9B & 1.4$\times$ \\
		BoTNet-S1-128~\cite{bot} & 83.5\% & 96.5\% & - & 75.1M & 256$^2$ & 19.3B & 2.1$\times$ \\
		EfficientNetV2-S~\cite{efficientnetv2} & 83.9\% & - & - & 22M & 224$^2$ & 8.8B & 1$\times$ \\
		EfficientNet-B6~\cite{efficientnet} & 84.0\% & 96.8\% & 96.9 & 43M & 528$^2$ & 19.2B & 2.0$\times$ \\
		\bf CMT-B  & \bf 84.5\% & \bf 96.9\% & 285.4 & 45.7M & 256$^2$ & \bf 9.3B & \bf 1$\times$ \\
		\midrule
    	EfficientNet-B7~\cite{efficientnet} & 84.3\% & 97.0\% & 55.1 & 66M & 600$^2$ & 37B & 1.9$\times$ \\
    	\bf CMT-L & \bf 84.8\% & \bf 97.1\% & 150.4 & 74.7M & 288$^2$ & \bf 19.5B & \bf 1$\times$ \\
		\bottomrule[0.15em]
	\end{tabular}
	\vspace{-0.2cm}
	\label{tab:imagenet}
\end{table*}

\section{Experiments}

In this section, we investigate the effectiveness of CMT architecture by conducting experiments on several tasks including image classification, object detection, and instance segmentation. We first compare the proposed CMT with previous state-of-the-art models on above tasks, and then ablate the important elements of CMT.

\subsection{ImageNet Classification}
\noindent\textbf{Experimental Settings.} ImageNet~\cite{imagenet} is a image classification benchmark which contains 1.28M training images and 50K validation images of 1000 classes. For fair comparisons with recent works, we adopt the same training and augmentation strategy as that in DeiT~\cite{deit}, \ie, models are trained for 300 epochs (800 for CMT-Ti that requires more epochs to converge) using the AdamW~\cite{adamw} optimizer. All models are trained on 8 NVIDIA Tesla V100 GPUs.

\vspace{0.1cm}
\noindent\textbf{Results of CMT.} Table~\ref{tab:imagenet} shows the performances of the proposed CMTs that are scaled from the CMT-S according to E.q.~\ref{eq:scaling}. Our models achieve better accuracy with fewer parameters and FLOPs compared to other convolution-based and transformer-based counterparts. In particular, our CMT-S achieves 83.5\% top-1 accuracy with 4.0B FLOPs, which is 3.7\% higher than the baseline model DeiT-S~\cite{deit} and 2.0\% higher than CPVT~\cite{cpvt}, indicating the benefit of CMT block for capturing both local and global information. Note that all previous transformer-based models are still inferior to EfficientNet~\cite{efficientnet} which is obtained via a thorough architecture search, however, our CMT-S is 0.6\% higher than EfficientNet-B4 with less computational cost, which demonstrates the efficacy of the proposed hybrid structure and show strong potential for further improvement. We also plot the accuracy-FLOPs curve in Figure~\ref{fig:imagenet-coco}(a) to have an intuitive comparison between these models. We can see that CMTs consistently outperform other models by a large margin.

\begin{wraptable}{h}{4.5cm}
	\centering
	\vskip -0.2in
	\caption{\small\textbf{Ablation study of stage-wise architecture on ImageNet.}}
	\vspace{-0.3cm}
	\label{tab:abl-stage}
	\footnotesize 
	\renewcommand\tabcolsep{2.0pt}
	\begin{tabular}{lccc}
		\toprule
		Model & Params & FLOPs & Top-1 \\
		\hline
		DeiT-S & 22M & 4.6B & 79.8\% \\
		DeiT-S-4Stage & 25M & 3.7B & 81.4\% \\
		\bottomrule
	\end{tabular}
	\vskip -0.1in
\end{wraptable}

\subsection{Ablation Study}
\noindent\textbf{Stage-wise architecture.} Transformer-based ViT/DeiT can only generate single-scale feature map, losing a lot of multi-scale information crucial for dense prediction tasks. We change the columnar DeiT-S to hierarchical DeiT-S-4Stage, which has 4 stages like CMT-S in Table~\ref{tab:arch}, but maintains the original FFN. We also change MHSA to LMHSA to reduce computational cost. As shown in Table~\ref{tab:abl-stage}, DeiT-S-4Stage outperforms DeiT-S by 1.6\% with less FLOPs, demonstrating that the widely-adopted stage-wise design in CNNs is a better choice for promoting transformer-based architecture.

\begin{table}[h]
	\centering
	\caption{\small\textbf{Ablation study of the scaling strategy.}}
	\vspace{-0.3cm}
	\label{tab:scaling}
	\footnotesize 
	\renewcommand\tabcolsep{2.0pt}
	\begin{tabular}{l|cc|cc}
		\toprule
		Model (based on CMT-S) & FLOPs & Top-1 & FLOPs & Top-1 \\
		\hline
		Scale: $\alpha$=2.2 (depth only) & 1.7B ($\phi$=-1) & 80.8\% & 8.6B($\phi$=1) & 83.4\% \\
		Scale: $\beta$=1.6 (dimension only) & 1.7B ($\phi$=-1) & 81.3\% & 9.3B($\phi$=1) & 83.8\%  \\
		Scale: $\alpha$=1.3, $\beta$=1.3, $\gamma$=1.15 & 1.5B ($\phi$=-1) & 81.8\% & 9.3B($\phi$=1) & 84.5\% \\
		\bottomrule
	\end{tabular}
	\vskip -0.25in
\end{table}

\vspace{0.1cm}
\noindent\textbf{CMT block.} Ablations on different modules in CMT are shown in Table~\ref{tab:abl-cmt-block}. DeiT-S-4Stage has 4 patch embedding layers (the first is a 4$\times$4 convolution with stride 4). ``+ Stem'' indicates that we add the CMT stem into the network and replace the first patch embedding layer with a 2$\times$2 convolution with stride 2. The improvement shows the benefit of the convolution-based stem. Besides, the proposed LPU and IRFFN can further boost the network by 0.8\% and 0.6\%, respectively. It is worth noticing that the shortcut connections in LPU and IRFFN are also crucial for the final performance.

\begin{wraptable}{h}{4.5cm}
	\centering
	\vskip -0.2in
	\caption{\small\textbf{Ablations of CMT block.}}
	\vspace{-0.3cm}
	\label{tab:abl-cmt-block}
	\footnotesize 
	\renewcommand\tabcolsep{2.0pt}
	\begin{tabular}{lccc}
		\toprule
		Model & Params & FLOPs & Top-1 \\
		\hline
		DeiT-S-4Stage & 25M & 3.7B & 81.4\% \\
		+ Stem & 25M & 3.9B & 81.9\% \\
		\hline
		+ LPU & 25M & 3.9B & 82.7\% \\
		w/o shortcut & 25M & 3.9B & 82.0\% \\
		\hline
		+ IRFFN & 25M & 3.9B & 83.3\% \\
		w/o shortcut & 25M & 3.9B & 82.5\% \\
		\hline
		+ Projection & 25M & 4.0B & 83.5\% \\
		\bottomrule
	\end{tabular}
	\vskip -0.1in
\end{wraptable}

\vspace{0.1cm}
\noindent\textbf{Normalization function.} Transformer-based models usually use LN~\cite{ln} inherited from NLP. However, convolution-based models usually utilize batch normalization (BN)~\cite{bn} to stabilize the training. CMT maintains the LN before LMHSA and IRFFN, and inserts BN after the convolutional layer. If all LNs are replaced by BNs, the model cannot converge during training. If all BNs are replaced by LNs, the performance of CMT-S drops to 83.0\%, indicating that proper application of normalization functions can improve the final performance.

\vspace{0.1cm}
\noindent\textbf{Scaling strategy.} Table~\ref{tab:scaling} shows the ImageNet results of CMT architecture under different scaling strategies. Unidimensional scaling strategies are significantly inferior to the proposed compound scaling strategy, especially for depth-only scaling strategy which leads to a even worse result of $83.4\%$ against $83.8\%$ of the original CMT-S, when the network is scaled up.

\begin{table*}[t]
	\setlength\tabcolsep{5.0pt}
	\caption{\small{\textbf{Object detection results on COCO val2017.} All models use RetinaNet~\cite{focal} as basic framework and are trained in ``1x'' schedule. FLOPs are calculated on 1280$\times$800 input. $\dagger$ means the results are from \cite{twins}.}}
	\vspace{-0.3cm}
	\label{tab:coco-retina}
	\centering
	\small 
	\begin{tabular}{l||cc||c||cc||ccc}
		\toprule
		Backbone & \# Params & \# FLOPs & mAP & $\rm {AP}_{50}$ & $\rm {AP}_{75}$ & $\rm {AP}_{S}$ & $\rm {AP}_{M}$ & $\rm {AP}_{L}$ \\
		\midrule
		ConT-M~\cite{cont} & 27.0M & 217B & 37.9 & 58.1 & 40.2 & 23.0 & 40.6 & 50.4 \\
		ResNet-101~\cite{resnet} & 56.7M & 315B & 38.5 & 57.6 & 41.0 & 21.7 & 42.8 & 50.4 \\
		RelationNet++~\cite{relationnet++} & 39.0M & 266B & 39.4 & 58.2 & 42.5 & - & - & - \\
		ResNeXt-101-32x4d~\cite{resnext} & 56.4M & 319B & 39.9 & 59.6 & 42.7 & 22.3 & 44.2 & 52.5 \\
		PVT-S~\cite{pvt} & 34.2M & 226B & 40.4 & 61.3 & 43.0 & 25.0 & 42.9 & 55.7 \\
		Swin-T$^{\dagger}$~\cite{swin} & 38.5M & 245B & 41.5 & 62.1 & 44.2 & 25.1 & 44.9 & 55.5 \\
		Twins-SVT-S~\cite{twins} & 34.3M & 209B & 42.3 & 63.4 & 45.2 & 26.0 & 45.5 & 56.5 \\
		Twins-PCPVT-S~\cite{twins} & 34.4M & 226B & 43.0 & 64.1 & 46.0 & 27.5 & 46.3 & 57.3 \\
		\textbf{CMT-S} (ours) & 44.3M & 231B & \bf 44.3 & \bf 65.5 & \bf 47.5 & \bf 27.1 & \bf 48.3 & \bf 59.1 \\
		\bottomrule
	\end{tabular}
	\vspace{-0.2cm}
\end{table*}

\begin{table*}[t]
	\setlength\tabcolsep{4.0pt}
	\caption{\small{\textbf{Instance segmentation results on COCO val2017.} All models use Mask R-CNN~\cite{maskrcnn} as basic framework and are trained in ``1x'' schedule. FLOPs are calculated on 1280$\times$800 input. $\dagger$ means the results are from \cite{twins}.}}
	\vspace{-0.3cm}
	\label{tab:coco-mask}
	\centering
	\small 
	\begin{tabular}{l||cc||c||cc||c||cc}
		\toprule
		Backbone & \# Params & \# FLOPs & AP$^{\rm box}$ & AP$_{50}^{\rm box}$ &AP$_{75}^{\rm box}$ &AP$^{\rm mask}$ &AP$_{50}^{\rm mask}$ &AP$_{75}^{\rm mask}$ \\
		\midrule
		ResNet-101~\cite{resnet} & 63.2M & 336B & 40.0 & 60.5 & 44.0 & 36.1 & 57.5 & 38.6 \\
		PVT-S~\cite{pvt} & 44.1M & 245B & 40.4 & 62.9 & 43.8 & 37.8 & 60.1 & 40.3 \\	ResNeXt-101-32x4d~\cite{resnext} & 62.8M & 340B & 41.9 & 62.5 & 45.9 & 37.5 & 59.4 & 40.2 \\
		Swin-T$^{\dagger}$~\cite{swin} & 47.8M & 264B & 42.2 & 64.6 & 46.2 & 39.1 & 61.6 & 42.0 \\
		Twins-SVT-S~\cite{twins} & 44.0M & 228B & 42.7 & 65.6 & 46.7 & 39.6 & 62.5 & 42.6 \\
		Twins-PCPVT-S~\cite{twins} & 44.3M & 245B & 42.9 & 65.8 & 47.1 & 40.0 & 62.7 & 42.9 \\
		\textbf{CMT-S} (ours) & 44.5M & 249B & \bf 44.6 & \bf 66.8 & \bf 48.9 & \bf 40.7 & \bf 63.9 & \bf 43.4 \\
		\bottomrule
	\end{tabular}
	\vspace{-0.2cm}
\end{table*}

\begin{table*}[!h]
	\caption{\textbf{Transfer Learning Results.} Models are fine-tuned with the ImageNet pretrained checkpoints. $\dagger$ means the results are from \cite{better}.}
	\vspace{-0.3cm}
	\centering
	\small
	\renewcommand\tabcolsep{5.0pt}
	\begin{tabular}{l||cc||c||c||c||c||c}
		\toprule [0.15em]
		Model & \# Params & \# FLOPs & CIFAR10 & CIFAR100 & Cars & Flowers & Pets \\
		\midrule [0.1em]
		ResNet-152$^{\dagger}$~\cite{resnet} & 58.1M & 11.3B & 97.9\% & 87.6\% & 92.0\% & 97.4\% & 94.5\% \\
		Inception-v4$^{\dagger}$~\cite{inceptionv4} & 41.1M & 16.1B & 97.9\% & 87.5\% & 93.3\% & 98.5\% & 93.7\% \\
		EfficientNet-B7$_{\uparrow600}$~\cite{efficientnet} & 64.0M & 37.2B & 98.9\% & \bf 91.7\% & \bf 94.7\% & 98.8\% & \bf 95.4\% \\
		ViT-B/16$_{\uparrow384}$~\cite{vit} & 85.8M & 17.6B & 98.1\% & 87.1\% & - & 89.5\% & 93.8\% \\
		DeiT-B~\cite{deit} & 85.8M & 17.6B & 99.1\% & 90.8\% & 92.1\% & 98.4\% & - \\
		CeiT-S$_{\uparrow384}$~\cite{ceit} & 24.2M & 12.9B & 99.1\% & 90.8\% & 94.1\% & 98.6\% & 94.9\% \\
		TNT-S$_{\uparrow384}$~\cite{tnt} & \bf 23.8M & 17.3B & 98.7\% & 90.1\% & - & \bf 98.8\% & 94.7\% \\
		\textbf{CMT-S} (ours)  & 25.1M & \bf 4.04B & \bf 99.2\% & \bf 91.7\% & 94.4\% & 98.7\% & 95.2\% \\
		\bottomrule[0.15em]
	\end{tabular}
	\vspace{-0.4cm}
	\label{tab:transfer}
\end{table*}

\subsection{Transfer Learning}

\subsubsection{Object Detection and Instance Segmentation}
\noindent\textbf{Experimental Settings.} The experiments are conducted on COCO~\cite{coco}, which contains 118K training images and 5K validation images of 80 classes. We evaluate the proposed CMT-S using two typical framework: RetinaNet~\cite{focal} and Mask R-CNN~\cite{maskrcnn} for object detection and instance segmentation, respectively. Specifically, we replace the original backbone with our CMT-S to build new detectors. All models are trained under standard single-scale and ``1x'' schedule (12 epochs) following PVT~\cite{pvt}.

\vspace{0.1cm}
\noindent\textbf{Results of CMT.}
We report the performance comparison results of object detection task and instance segmentation task in Table~\ref{tab:coco-retina} and Table~\ref{tab:coco-mask} respectively.
For object detection with RetinaNet as basic framework, CMT-S outperforms Twins-PCPVT-S~\cite{twins} with 1.3\% mAP and Twins-SVT-S~\cite{twins} with 2.0\% mAP. For instance segmentation with Mask R-CNN as basic framework, CMT-S surpasses Twins-PCPVT-S~\cite{twins} with 1.7\% AP and Twins-SVT-S~\cite{twins} with 1.9\% AP.
We also report the inference speed on COCO val2017 with 1280×800 input, CMT-S based RetinaNet and Mask R-CNN achieve 14.8 FPS and 11.2 FPS, respectively.

\vspace{-0.2cm}
\subsubsection{Other Vision Tasks}
\vspace{-0.1cm}
We also evaluate the proposed CMT on five commonly used transfer learning datasets, including CIFAR10~\cite{cifar}, CIFAR100~\cite{cifar}, Standford Cars~\cite{cars}, Flowers~\cite{flowers}, and Oxford-IIIT Pets~\cite{pets} (see Appendix for more details). We fine-tune the ImageNet pretrained models on new datasets following~\cite{efficientnet,tnt}. Table~\ref{tab:transfer} shows the corresponding results. CMT-S outperforms other transformer-based models in all datasets with less FLOPs, and achieves comparable performance against EfficientNet-B7~\cite{efficientnet} with 9x less FLOPs, which demonstrates the superiority of CMT architecture.

\section{Conclusion}
This paper proposes a novel hybrid architecture named CMT for visual recognition and other downstream computer vision tasks such as object detection and instance segmentation, and addresses the limitations of utilizing transformers in a brutal force manner in the field of computer vision.
The proposed CMT architectures take advantages of both CNNs and transformers to capture local and global information, promoting the representation ability of the network.
In addition, a scaling strategy is proposed to generate a family of CMT variants for different resource constraints.
Extensive experiments on ImageNet and other downstream vision tasks demonstrate the effectiveness and superiority of the proposed CMT architecture.

\noindent \textbf{Acknowledgment} Chang Xu was supported by the Australian Research Council under Project DP210101859 and the University of Sydney SOAR Prize.

{\small
	\bibliographystyle{ieee_fullname}
	\bibliography{egbib}

\begin{thebibliography}{10}\itemsep=-1pt

\bibitem{ln}
Jimmy~Lei Ba, Jamie~Ryan Kiros, and Geoffrey~E Hinton.
\newblock Layer normalization.
\newblock {\em arXiv preprint arXiv:1607.06450}, 2016.

\bibitem{detr}
Nicolas Carion, Francisco Massa, Gabriel Synnaeve, Nicolas Usunier, Alexander
  Kirillov, and Sergey Zagoruyko.
\newblock End-to-end object detection with transformers.
\newblock In {\em European Conference on Computer Vision}, 2020.

\bibitem{ipt}
Hanting Chen, Yunhe Wang, Tianyu Guo, Chang Xu, Yiping Deng, Zhenhua Liu, Siwei
  Ma, Chunjing Xu, Chao Xu, and Wen Gao.
\newblock Pre-trained image processing transformer.
\newblock {\em arXiv preprint arXiv:2012.00364}, 2020.

\bibitem{relationnet++}
Cheng Chi, Fangyun Wei, and Han Hu.
\newblock Relationnet++: Bridging visual representations for object detection
  via transformer decoder.
\newblock {\em arXiv preprint arXiv:2010.15831}, 2020.

\bibitem{twins}
Xiangxiang Chu, Zhi Tian, Yuqing Wang, Bo Zhang, Haibing Ren, Xiaolin Wei,
  Huaxia Xia, and Chunhua Shen.
\newblock Twins: Revisiting spatial attention design in vision transformers.
\newblock {\em arXiv preprint arXiv:2104.13840}, 2021.

\bibitem{cpvt}
Xiangxiang Chu, Bo Zhang, Zhi Tian, Xiaolin Wei, and Huaxia Xia.
\newblock Conditional positional encodings for vision transformers.
\newblock {\em arXiv preprint arXiv:2102.10882}, 2021.

\bibitem{cityscapes}
Marius Cordts, Mohamed Omran, Sebastian Ramos, Timo Rehfeld, Markus Enzweiler,
  Rodrigo Benenson, Uwe Franke, Stefan Roth, and Bernt Schiele.
\newblock The cityscapes dataset for semantic urban scene understanding.
\newblock In {\em Proceedings of the IEEE conference on computer vision and
  pattern recognition}, 2016.

\bibitem{imagenet}
Jia Deng, Wei Dong, Richard Socher, Li-Jia Li, Kai Li, and Li Fei-Fei.
\newblock Imagenet: A large-scale hierarchical image database.
\newblock In {\em 2009 IEEE conference on computer vision and pattern
  recognition}, 2009.

\bibitem{bert}
Jacob Devlin, Ming-Wei Chang, Kenton Lee, and Kristina Toutanova.
\newblock Bert: Pre-training of deep bidirectional transformers for language
  understanding.
\newblock In {\em arXiv preprint arXiv:1810.04805}, 2018.

\bibitem{vit}
Alexey Dosovitskiy, Lucas Beyer, Alexander Kolesnikov, Dirk Weissenborn,
  Xiaohua Zhai, Thomas Unterthiner, Mostafa Dehghani, Matthias Minderer, Georg
  Heigold, Sylvain Gelly, et~al.
\newblock An image is worth 16x16 words: Transformers for image recognition at
  scale.
\newblock {\em arXiv preprint arXiv:2010.11929}, 2020.

\bibitem{retrieval}
Alaaeldin El-Nouby, Natalia Neverova, Ivan Laptev, and Herv{\'e} J{\'e}gou.
\newblock Training vision transformers for image retrieval.
\newblock {\em arXiv preprint arXiv:2102.05644}, 2021.

\bibitem{hire}
Jianyuan Guo, Yehui Tang, Kai Han, Xinghao Chen, Han Wu, Chao Xu, Chang Xu, and
  Yunhe Wang.
\newblock Hire-mlp: Vision mlp via hierarchical rearrangement.
\newblock {\em arXiv preprint arXiv:2108.13341}, 2021.

\bibitem{ghostnet}
Kai Han, Yunhe Wang, Qi Tian, Jianyuan Guo, Chunjing Xu, and Chang Xu.
\newblock Ghostnet: More features from cheap operations.
\newblock In {\em Proceedings of the IEEE/CVF Conference on Computer Vision and
  Pattern Recognition}, 2020.

\bibitem{tnt}
Kai Han, An Xiao, Enhua Wu, Jianyuan Guo, Chunjing Xu, and Yunhe Wang.
\newblock Transformer in transformer.
\newblock {\em arXiv preprint arXiv:2103.00112}, 2021.

\bibitem{maskrcnn}
Kaiming He, Georgia Gkioxari, Piotr Doll{\'a}r, and Ross Girshick.
\newblock Mask r-cnn.
\newblock In {\em Proceedings of the IEEE international conference on computer
  vision}, 2017.

\bibitem{resnet}
Kaiming He, Xiangyu Zhang, Shaoqing Ren, and Jian Sun.
\newblock Deep residual learning for image recognition.
\newblock In {\em Proceedings of the IEEE conference on computer vision and
  pattern recognition}, 2016.

\bibitem{he2019bag}
Tong He, Zhi Zhang, Hang Zhang, Zhongyue Zhang, Junyuan Xie, and Mu Li.
\newblock Bag of tricks for image classification with convolutional neural
  networks.
\newblock In {\em Proceedings of the IEEE/CVF Conference on Computer Vision and
  Pattern Recognition}, 2019.

\bibitem{gelu}
Dan Hendrycks and Kevin Gimpel.
\newblock Gaussian error linear units (gelus).
\newblock {\em arXiv preprint arXiv:1606.08415}, 2016.

\bibitem{mobilenetv3}
Andrew Howard, Mark Sandler, Grace Chu, Liang-Chieh Chen, Bo Chen, Mingxing
  Tan, Weijun Wang, Yukun Zhu, Ruoming Pang, Vijay Vasudevan, et~al.
\newblock Searching for mobilenetv3.
\newblock In {\em Proceedings of the IEEE/CVF International Conference on
  Computer Vision}, 2019.

\bibitem{genet}
Jie Hu, Li Shen, Samuel Albanie, Gang Sun, and Andrea Vedaldi.
\newblock Gather-excite: Exploiting feature context in convolutional neural
  networks.
\newblock {\em arXiv preprint arXiv:1810.12348}, 2018.

\bibitem{senet}
Jie Hu, Li Shen, and Gang Sun.
\newblock Squeeze-and-excitation networks.
\newblock In {\em Proceedings of the IEEE conference on computer vision and
  pattern recognition}, 2018.

\bibitem{densenet}
Gao Huang, Zhuang Liu, Laurens Van Der~Maaten, and Kilian~Q Weinberger.
\newblock Densely connected convolutional networks.
\newblock In {\em Proceedings of the IEEE conference on computer vision and
  pattern recognition}, 2017.

\bibitem{mindspore}
Mindspore. Huawei.
\newblock \url{https://www.mindspore.cn/}.
\newblock 2020.

\bibitem{squeezenet}
Forrest~N Iandola, Song Han, Matthew~W Moskewicz, Khalid Ashraf, William~J
  Dally, and Kurt Keutzer.
\newblock Squeezenet: Alexnet-level accuracy with 50x fewer parameters and< 0.5
  mb model size.
\newblock {\em arXiv preprint arXiv:1602.07360}, 2016.

\bibitem{bn}
Sergey Ioffe and Christian Szegedy.
\newblock Batch normalization: Accelerating deep network training by reducing
  internal covariate shift.
\newblock In {\em International conference on machine learning}, 2015.

\bibitem{zero-padding}
Md~Amirul Islam, Sen Jia, and Neil~DB Bruce.
\newblock How much position information do convolutional neural networks
  encode?
\newblock {\em arXiv preprint arXiv:2001.08248}, 2020.

\bibitem{translation-invariance}
Osman~Semih Kayhan and Jan C~van Gemert.
\newblock On translation invariance in cnns: Convolutional layers can exploit
  absolute spatial location.
\newblock In {\em Proceedings of the IEEE/CVF Conference on Computer Vision and
  Pattern Recognition}, 2020.

\bibitem{better}
Simon Kornblith, Jonathon Shlens, and Quoc~V Le.
\newblock Do better imagenet models transfer better?
\newblock In {\em Proceedings of the IEEE/CVF Conference on Computer Vision and
  Pattern Recognition}, 2019.

\bibitem{cars}
Jonathan Krause, Michael Stark, Jia Deng, and Li Fei-Fei.
\newblock 3d object representations for fine-grained categorization.
\newblock In {\em Proceedings of the IEEE international conference on computer
  vision workshops}, 2013.

\bibitem{cifar}
Alex Krizhevsky, Geoffrey Hinton, et~al.
\newblock Learning multiple layers of features from tiny images.
\newblock 2009.

\bibitem{alexnet}
Alex Krizhevsky, Ilya Sutskever, and Geoffrey~E Hinton.
\newblock Imagenet classification with deep convolutional neural networks.
\newblock {\em Advances in neural information processing systems}, 2012.

\bibitem{lenet}
Yann LeCun, L{\'e}on Bottou, Yoshua Bengio, and Patrick Haffner.
\newblock Gradient-based learning applied to document recognition.
\newblock {\em Proceedings of the IEEE}, 1998.

\bibitem{li2022brain}
Wenshuo Li, Hanting Chen, Jianyuan Guo, Ziyang Zhang, and Yunhe Wang.
\newblock Brain-inspired multilayer perceptron with spiking neurons.
\newblock {\em arXiv preprint arXiv:2203.14679}, 2022.

\bibitem{focal}
Tsung-Yi Lin, Priya Goyal, Ross Girshick, Kaiming He, and Piotr Doll{\'a}r.
\newblock Focal loss for dense object detection.
\newblock In {\em Proceedings of the IEEE international conference on computer
  vision}, 2017.

\bibitem{coco}
Tsung-Yi Lin, Michael Maire, Serge Belongie, James Hays, Pietro Perona, Deva
  Ramanan, Piotr Doll{\'a}r, and C~Lawrence Zitnick.
\newblock Microsoft coco: Common objects in context.
\newblock In {\em European conference on computer vision}. Springer.

\bibitem{swin}
Ze Liu, Yutong Lin, Yue Cao, Han Hu, Yixuan Wei, Zheng Zhang, Stephen Lin, and
  Baining Guo.
\newblock Swin transformer: Hierarchical vision transformer using shifted
  windows.
\newblock {\em arXiv preprint arXiv:2103.14030}, 2021.

\bibitem{adamw}
Ilya Loshchilov and Frank Hutter.
\newblock Decoupled weight decay regularization.
\newblock {\em arXiv preprint arXiv:1711.05101}, 2017.

\bibitem{local-scale}
David~G Lowe.
\newblock Object recognition from local scale-invariant features.
\newblock In {\em Proceedings of the seventh IEEE international conference on
  computer vision}, 1999.

\bibitem{shufflenetv2}
Ningning Ma, Xiangyu Zhang, Hai-Tao Zheng, and Jian Sun.
\newblock Shufflenet v2: Practical guidelines for efficient cnn architecture
  design.
\newblock In {\em Proceedings of the European conference on computer vision
  (ECCV)}, 2018.

\bibitem{flowers}
Maria-Elena Nilsback and Andrew Zisserman.
\newblock Automated flower classification over a large number of classes.
\newblock In {\em 2008 Sixth Indian Conference on Computer Vision, Graphics \&
  Image Processing}, 2008.

\bibitem{bam}
Jongchan Park, Sanghyun Woo, Joon-Young Lee, and In~So Kweon.
\newblock Bam: Bottleneck attention module.
\newblock {\em arXiv preprint arXiv:1807.06514}, 2018.

\bibitem{pets}
Omkar~M Parkhi, Andrea Vedaldi, Andrew Zisserman, and CV Jawahar.
\newblock Cats and dogs.
\newblock In {\em 2012 IEEE conference on computer vision and pattern
  recognition}, 2012.

\bibitem{pytorch}
Adam Paszke, Sam Gross, Francisco Massa, Adam Lerer, James Bradbury, Gregory
  Chanan, Trevor Killeen, Zeming Lin, Natalia Gimelshein, Luca Antiga, et~al.
\newblock Pytorch: An imperative style, high-performance deep learning library.
\newblock {\em arXiv preprint arXiv:1912.01703}, 2019.

\bibitem{regnety}
Ilija Radosavovic, Raj~Prateek Kosaraju, Ross Girshick, Kaiming He, and Piotr
  Doll{\'a}r.
\newblock Designing network design spaces.
\newblock In {\em Proceedings of the IEEE/CVF Conference on Computer Vision and
  Pattern Recognition}, 2020.

\bibitem{sasa}
Prajit Ramachandran, Niki Parmar, Ashish Vaswani, Irwan Bello, Anselm Levskaya,
  and Jonathon Shlens.
\newblock Stand-alone self-attention in vision models.
\newblock {\em arXiv preprint arXiv:1906.05909}, 2019.

\bibitem{mobilenetv2}
Mark Sandler, Andrew Howard, Menglong Zhu, Andrey Zhmoginov, and Liang-Chieh
  Chen.
\newblock Mobilenetv2: Inverted residuals and linear bottlenecks.
\newblock In {\em Proceedings of the IEEE conference on computer vision and
  pattern recognition}, 2018.

\bibitem{vggnet}
Karen Simonyan and Andrew Zisserman.
\newblock Very deep convolutional networks for large-scale image recognition.
\newblock {\em arXiv preprint arXiv:1409.1556}, 2014.

\bibitem{bot}
Aravind Srinivas, Tsung-Yi Lin, Niki Parmar, Jonathon Shlens, Pieter Abbeel,
  and Ashish Vaswani.
\newblock Bottleneck transformers for visual recognition.
\newblock {\em arXiv preprint arXiv:2101.11605}, 2021.

\bibitem{jft}
Chen Sun, Abhinav Shrivastava, Saurabh Singh, and Abhinav Gupta.
\newblock Revisiting unreasonable effectiveness of data in deep learning era.
\newblock In {\em Proceedings of the IEEE international conference on computer
  vision}, 2017.

\bibitem{inceptionv4}
Christian Szegedy, Sergey Ioffe, Vincent Vanhoucke, and Alexander Alemi.
\newblock Inception-v4, inception-resnet and the impact of residual connections
  on learning.
\newblock In {\em Proceedings of the AAAI Conference on Artificial
  Intelligence}, 2017.

\bibitem{googlenet}
Christian Szegedy, Wei Liu, Yangqing Jia, Pierre Sermanet, Scott Reed, Dragomir
  Anguelov, Dumitru Erhan, Vincent Vanhoucke, and Andrew Rabinovich.
\newblock Going deeper with convolutions.
\newblock In {\em Proceedings of the IEEE conference on computer vision and
  pattern recognition}, 2015.

\bibitem{inceptionv3}
Christian Szegedy, Vincent Vanhoucke, Sergey Ioffe, Jon Shlens, and Zbigniew
  Wojna.
\newblock Rethinking the inception architecture for computer vision.
\newblock In {\em Proceedings of the IEEE conference on computer vision and
  pattern recognition}, 2016.

\bibitem{efficientnet}
Mingxing Tan and Quoc Le.
\newblock Efficientnet: Rethinking model scaling for convolutional neural
  networks.
\newblock In {\em International Conference on Machine Learning}, 2019.

\bibitem{efficientnetv2}
Mingxing Tan and Quoc Le.
\newblock Efficientnetv2: Smaller models and faster training.
\newblock In {\em International Conference on Machine Learning}, 2021.

\bibitem{wave}
Yehui Tang, Kai Han, Jianyuan Guo, Chang Xu, Yanxi Li, Chao Xu, and Yunhe Wang.
\newblock An image patch is a wave: Phase-aware vision mlp.
\newblock {\em arXiv preprint arXiv:2111.12294}, 2021.

\bibitem{tang2021patch}
Yehui Tang, Kai Han, Yunhe Wang, Chang Xu, Jianyuan Guo, Chao Xu, and Dacheng
  Tao.
\newblock Patch slimming for efficient vision transformers.
\newblock {\em arXiv preprint arXiv:2106.02852}, 2021.

\bibitem{deit}
Hugo Touvron, Matthieu Cord, Matthijs Douze, Francisco Massa, Alexandre
  Sablayrolles, and Herv{\'e} J{\'e}gou.
\newblock Training data-efficient image transformers \& distillation through
  attention.
\newblock {\em arXiv preprint arXiv:2012.12877}, 2020.

\bibitem{attention}
Ashish Vaswani, Noam Shazeer, Niki Parmar, Jakob Uszkoreit, Llion Jones,
  Aidan~N Gomez, Lukasz Kaiser, and Illia Polosukhin.
\newblock Attention is all you need.
\newblock In {\em arXiv preprint arXiv:1706.03762}, 2017.

\bibitem{wang2017residual}
Fei Wang, Mengqing Jiang, Chen Qian, Shuo Yang, Cheng Li, Honggang Zhang,
  Xiaogang Wang, and Xiaoou Tang.
\newblock Residual attention network for image classification.
\newblock In {\em Proceedings of the IEEE conference on computer vision and
  pattern recognition}, 2017.

\bibitem{pvt}
Wenhai Wang, Enze Xie, Xiang Li, Deng-Ping Fan, Kaitao Song, Ding Liang, Tong
  Lu, Ping Luo, and Ling Shao.
\newblock Pyramid vision transformer: A versatile backbone for dense prediction
  without convolutions.
\newblock {\em arXiv preprint arXiv:2102.12122}, 2021.

\bibitem{nlnet}
Xiaolong Wang, Ross Girshick, Abhinav Gupta, and Kaiming He.
\newblock Non-local neural networks.
\newblock In {\em Proceedings of the IEEE conference on computer vision and
  pattern recognition}, 2018.

\bibitem{cbam}
Sanghyun Woo, Jongchan Park, Joon-Young Lee, and In~So Kweon.
\newblock Cbam: Convolutional block attention module.
\newblock In {\em Proceedings of the European conference on computer vision
  (ECCV)}, 2018.

\bibitem{cvt}
Haiping Wu, Bin Xiao, Noel Codella, Mengchen Liu, Xiyang Dai, Lu Yuan, and Lei
  Zhang.
\newblock Cvt: Introducing convolutions to vision transformers.
\newblock {\em arXiv preprint arXiv:2103.15808}, 2021.

\bibitem{resnext}
Saining Xie, Ross Girshick, Piotr Doll{\'a}r, Zhuowen Tu, and Kaiming He.
\newblock Aggregated residual transformations for deep neural networks.
\newblock In {\em Proceedings of the IEEE conference on computer vision and
  pattern recognition}, 2017.

\bibitem{coat}
Weijian Xu, Yifan Xu, Tyler Chang, and Zhuowen Tu.
\newblock Co-scale conv-attentional image transformers.
\newblock {\em arXiv preprint arXiv:2104.06399}, 2021.

\bibitem{cont}
Haotian Yan, Zhe Li, Weijian Li, Changhu Wang, Ming Wu, and Chuang Zhang.
\newblock Contnet: Why not use convolution and transformer at the same time?
\newblock {\em arXiv preprint arXiv:2104.13497}, 2021.

\bibitem{ceit}
Kun Yuan, Shaopeng Guo, Ziwei Liu, Aojun Zhou, Fengwei Yu, and Wei Wu.
\newblock Incorporating convolution designs into visual transformers.
\newblock {\em arXiv preprint arXiv:2103.11816}, 2021.

\bibitem{t2t}
Li Yuan, Yunpeng Chen, Tao Wang, Weihao Yu, Yujun Shi, Francis~EH Tay, Jiashi
  Feng, and Shuicheng Yan.
\newblock Tokens-to-token vit: Training vision transformers from scratch on
  imagenet.
\newblock {\em arXiv preprint arXiv:2101.11986}, 2021.

\bibitem{setr}
Sixiao Zheng, Jiachen Lu, Hengshuang Zhao, Xiatian Zhu, Zekun Luo, Yabiao Wang,
  Yanwei Fu, Jianfeng Feng, Tao Xiang, Philip~HS Torr, et~al.
\newblock Rethinking semantic segmentation from a sequence-to-sequence
  perspective with transformers.
\newblock {\em arXiv preprint arXiv:2012.15840}, 2020.

\bibitem{deformable-detr}
Xizhou Zhu, Weijie Su, Lewei Lu, Bin Li, Xiaogang Wang, and Jifeng Dai.
\newblock Deformable detr: Deformable transformers for end-to-end object
  detection.
\newblock {\em arXiv preprint arXiv:2010.04159}, 2020.

\end{thebibliography}
}

\end{document}